\UseRawInputEncoding
\documentclass[journal,twoside,web]{ieeecolor}
\usepackage{threeparttable}
\usepackage{tmi}
\usepackage{cite}
\usepackage{amsmath,amssymb,amsfonts}
\usepackage{algorithmic}
\usepackage{graphicx}
\usepackage{textcomp}
\usepackage{multirow}
\pdfoutput=1

\def\BibTeX{{\rm B\kern-.05em{\sc i\kern-.025em b}\kern-.08em
    T\kern-.1667em\lower.7ex\hbox{E}\kern-.125emX}}
\begin{document}
\title{Instrument-tissue Interaction Detection Framework for Surgical Video Understanding}
\author{Wenjun Lin, Yan Hu, Huazhu Fu, Mingming Yang, Chin-Boon Chng, Ryo Kawasaki, \\ Cheekong Chui, Jiang Liu, \IEEEmembership{Senior Member, IEEE}
\thanks{Wenjun Lin and Yan Hu are co-first authors.  CA: Yan Hu, Jiang Liu, and Cheekong Chui. Wenjun Lin, Chin-Boon Chng, and Cheekong Chui are with the Department of Mechanical Engineering, National University of Singapore; 
Wenjun Lin, Yan Hu, and Jiang Liu are with the Research Institute of Trustworthy Autonomous Systems and the Department of Computer Science and Engineering, Southern University of Science and Technology (e-mail: huy3@sustech.edu.cn); Huazhu Fu is with the Agency for Science, Technology, and Research; Ryo Kawasaki is with the Department of Informatics, Osaka University Graduate School of Medicine; Mingming Yang is with the Department of Ophthalmology, Shenzhen People’s Hospital.}
}

\maketitle

\begin{abstract}
Instrument-tissue interaction detection task, which helps understand surgical activities, is vital for constructing computer-assisted surgery systems but with many challenges. Firstly, most models represent instrument-tissue interaction in a coarse-grained way which only focuses on classification and lacks the ability to automatically detect instruments and tissues. Secondly, existing works do not fully consider relations between intra- and inter-frame of instruments and tissues. In the paper, we propose to represent instrument-tissue interaction as $\langle$instrument class, instrument bounding box, tissue class, tissue bounding box, action class$\rangle$ quintuple and present an Instrument-Tissue Interaction Detection Network (ITIDNet) to detect the quintuple for surgery videos understanding. Specifically, we propose a Snippet Consecutive Feature (SCF) Layer to enhance features by modeling relationships of proposals in the current frame using global context information in the video snippet. We also propose a Spatial Corresponding Attention (SCA) Layer to incorporate features of proposals between adjacent frames through spatial encoding. To reason relationships between instruments and tissues, a Temporal Graph (TG) Layer is proposed with intra-frame connections to exploit relationships between instruments and tissues in the same frame and inter-frame connections to model the temporal information for the same instance. For evaluation, we build a cataract surgery video (PhacoQ) dataset and a cholecystectomy surgery video (CholecQ) dataset. Experimental results demonstrate the promising performance of our model, which outperforms other state-of-the-art models on both datasets.
\end{abstract}


\section{Introduction}
\label{sec:introduction}
\begin{figure}[ht]
\centerline{\includegraphics[width=\columnwidth]{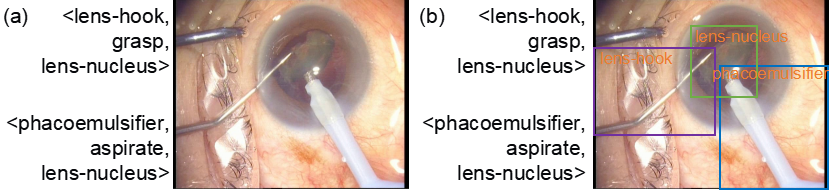}}
\caption{Examples of the existing instrument-tissue interaction recognition and our detection task. (a) Instrument-tissue interaction recognition task \cite{nwoye2020recognition}, represented as triplets. 
(b) Our instrument-tissue interaction detection task, represented as quintuples. 
}
\label{fig_example}
\end{figure}

\IEEEPARstart{C}{omputer} 
assisted surgery (CAS) has been a leading factor in the development of robotic surgery which assists doctors in performing surgeries based on their understanding of surgical scenes to improve surgical safety and quality \cite{maie2017surgical}. Instrument-tissue interaction detection with both category and location information to describe activities, as shown in Fig. \ref{fig_example} (b), is an essential prerequisite for constructing CAS systems \cite{lin2022instrument}. It provides a deep understanding of the environment in which surgical systems are working to provide assistance for performing autonomous tasks and making decisions autonomously. 
Incorporating surgical scene understanding capability in CAS systems provides possibilities for future semi-automated or fully automated operations \cite{feng2009application}.

We identify two main limitations in recent studies for identifying instrument-tissue interaction. Firstly, most recent works did not focus on the automatic detection of instruments and tissues. Some works \cite{islam2020learning}, \cite{xu2021learning}, \cite{seenivasan2022global} rely on the ground-truth instrument and tissue position for further instrument-tissue interaction inference. Some other works \cite{nwoye2020recognition}, \cite{nwoye2022rendezvous}, \cite{li2022sirnet}, as shown in Fig. \ref{fig_example} (a), only focus on category information and did not provide location information of instruments and tissues. 
The location information is essential to instructing operations of surgical robots and monitoring surgical safety, such as alerting surgeons when a dangerous area is touched \cite{sarikaya2017detection}. However, recent studies did not pay much attention to obtaining the location of instruments and tissues.

Secondly, existing works only consider intra-frame relationships between instruments and tissues and did not fully exploit instrument-tissue relationships between intra- and inter-frame. Interaction prediction is quite challenging since there are intra-class variances for the same action verb with different instruments and limited inter-class variances between different actions with the same instruments. 
Some existing works \cite{islam2020learning}, \cite{xu2021learning}, \cite{seenivasan2022global}, \cite{lin2022instrument} consider instrument-tissue relationships in the current frame to improve the prediction accuracy of actions. However, these works lack modeling temporal relationships for action prediction while action prediction relies heavily on temporal information since actions take time to execute.  

In this paper, we propose to represent instrument-tissue interaction as quintuple $\langle$instrument class, instrument bounding box, tissue class, tissue bounding box, action class$\rangle$. A novel model named Instrument-Tissue Interaction Detection Network (ITIDNet) that detects instances (instruments and tissues) in the first stage and predicts interactions for instrument-tissue pairs in the second stage is presented for the instrument-tissue interaction detection task. In the instance detection stage, some improvements have been made to the basic object detection network to better detect instruments and tissues. 
Firstly, a Snippet Consecutive Feature Layer is proposed to combine global context information of the video snippet with regional visual features and enhance features by exploiting relationships between proposals in the same frame. 
When object appearance is insufficient for object detection due to occlusion or poor image quality, global context information is a good aid to improve performance. 
Secondly, a Spatial Corresponding Attention Layer is proposed to exploit relationships between proposals in adjacent frames with the assistance of spatial information. 
For the interaction prediction stage, a Temporal Graph Layer is proposed to model instrument-tissue relationships and predict actions for instrument-tissue pairs through a constructed interaction graph. In this layer, relationships of instruments and tissues in the same frame are modeled through intra-frame connections. Moreover, temporal information of the same instance in consecutive video frames is essential for action prediction since the implementation of an action takes a period of time. Therefore, relationships of the same instance in adjacent frames are modeled through inter-frame connections in this layer.

This work is extended based on our conference paper \cite{lin2022instrument} in both method and data. The proposed ITIDNet improves on QDNet \cite{lin2022instrument} by comprehensively considering inter- and intra-frame relationships in both instance detection and interaction prediction stages. A cataract surgery video (PhacoQ) Dataset is extended based on the Cataract dataset \cite{lin2022instrument} to include frames with no instrument-tissue interaction. Furthermore, a cholecystectomy surgery video (CholecQ) Dataset is created for further evaluation.
The code and the dataset can be found at the link \footnote{https://gaiakoen.github.io/yanhu/research/Surgical\_Scenarios\_Understanding/}. 
In conclusion, our contributions are summarized as follows:
\begin{itemize} 
\item [1.] We inherit the concept of instrument-tissue interaction which represents interaction as a quintuple $\langle$instrument class, instrument bounding box, tissue class, tissue bounding box, action class$\rangle$ to understand the surgical scene details.  
We build two surgery video datasets to detect the quintuple, including a cataract PhacoQ dataset and a cholecystectomy CholecQ dataset, which will be publicly available. 
\item [2.]
We propose a novel ITIDNet, which jointly considers intra- and inter-frame relationships for both instance detection and interaction prediction. The model improves instrument-tissue interaction detection results compared to the state-of-the-art methods on our built datasets.
\item [3.]
For the instance detection stage, we introduce a Snippet Consecutive Feature (SCF) Layer modeling instrument-tissue relationships in the same frame and a Spatial Corresponding Attention (SCA) Layer adopting instrument-tissue relation between adjacent frames to improve the detection accuracy of instruments and tissues.
\item [4.] 
For the interaction prediction stage, based on our constructed interaction graph, we propose a Temporal Graph Layer to further refine the relationship of instruments and tissues in the same frame and the temporal relation of the same instance in adjacent frames for action prediction.

\end{itemize}

\section{Related Work}
\subsection{Surgical Activity Recognition}
Most existing works for surgical scene understanding focus on coarse-grained activity descriptions. Previous works on surgical activity recognition mainly defined an activity as a sequence of actions or gestures \cite{lea2015an}, \cite{khatibi2020proposing}, \cite{lea2016temporal}.  
DiPietro et al. \cite{dipietro2016recognizing} represented surgical activity as low-level gestures and higher-level maneuvers. 
Bawa et al. \cite{bawa2021saras} used verb-target combined nouns to represent surgical activities and the tips of instruments to represent the location of interaction.  
The above works focus on action verbs, ignoring the subject or object of the action. Such surgical activity representations can not adequately describe the information in the surgical scene.

Fine-grained surgical instrument-tissue interaction recognition has received attention in recent years with the development of automatic surgery.
To gain a fuller understanding of surgical scenes, Nwoye et al. \cite{nwoye2020recognition} first proposed to directly predict $\langle$instrument, tissue, action$\rangle$ triplet labels to describe activities in surgical scenes in 2020. This is defined as instrument-tissue interaction recognition which is a multi-label classification task and they created the first laparoscopic surgical dataset for this task. In 2022, they presented an extended work in both methods and data for instrument-tissue interaction recognition \cite{nwoye2022rendezvous}. Furthermore, Li et al. \cite{li2022sirnet} presented a Transformer-like architecture named SIRNet for the surgical instrument-tissue interaction recognition task.

Some other works for instrument-tissue interaction recognition focus on relationship reasoning when the location and type of instrument and tissue are known. Islam et al. \cite{islam2020learning} developed a graph network-based method using Graph Parsing Neural Networks \cite{qi2018learning} with SageConv \cite{hamilton2017inductive} to infer the interaction type with knowing instrument and target positions and types. Xu et al. \cite{xu2021learning} developed a multi-layer transformer-based model with gradient reversal adversarial learning to model the relationship between instruments and tissues for surgical report generation. Seenivasan et al. \cite{seenivasan2022global} proposed a globally-reasoned multi-task surgical scene understanding model to perform instrument segmentation and recognize instrument-tissue interaction with ground-truth instrument and tissue position and type. 

The above-mentioned existing works in the surgical instrument-tissue interaction task only focus on category information and overlook the detection of specific locations of instruments and tissues. 
A detailed description of activities in the given surgical scene should include instruments, target tissues, and their locations and relations. However, most existing works involve only a portion of this combination, with only category information and no information on the precise location of instruments or tissues.

\subsection{Human-object Interaction Detection}
Human-object interaction (HOI) detection \cite{chao2015hico} with both category and location information to describe human activities is a fundamental problem for scene understanding in natural scenes. A HOI detector is to locate human-object pairs and classify their corresponding action. Algorithms in this field provide references for instrument-tissue interaction detection.

Most HOI methods share a serial architecture. In the first stage, humans and objects are detected using off-the-shelf object detection models first. Then in the second stage, human-object pairs, which are generated by matching humans and objects one by one based on regional features, are fed to a network to predict the action between them.  
In recent years, with the successful application of the Transformer \cite{vaswani2017attention} in the field of image processing, transformer-based methods are also starting to appear in the HOI detection field \cite{tamura2021qpic}, \cite{zou2021end}. However, it's difficult to train the model well because it is hard to generate a unified feature representation for two very different tasks.  
Recently, Zhang et al. \cite{zhang2022efficient} argued that the success of Transformer-inspired HOI detectors can largely be attributed to the representation power of transformers. They proposed a two-stage Unary-Pairwise Transformer which outperforms state-of-the-art models. This proves that improving the detection model in the first stage is also a new way for performance improvement in the HOI detection task. 

Surgical scenes differ from natural scenes in many ways. Firstly, the boundaries of most tissues are unclear such as transparent tissues in cataract surgeries. Camera movement and bleeding also obscure the field of view, resulting in many missed and false detections. Secondly, the subject of an action in HOI is always human, while a variety of instruments can be the subject of an action type in surgical scenes. For different kinds of subjects, there might be the same kind of verb description. Therefore, directly applying HOI detection methods to surgical scenes cannot achieve desired results.

\section{Datasets}
\begin{table*}\centering
\caption{Summary of existing surgical workflow datasets. }
\label{tab_dataset}
\begin{threeparttable}
\begin{tabular}{|p{57pt}|p{23pt}|p{27pt}|p{70pt}|p{52pt}|p{20pt}|p{20pt}|p{20pt}|p{20pt}|p{30pt}|}
\hline
Dataset & \#Videos & \#Frames & Annotation & Anatomy & \#Avt & \#Ins & \#Act & \#Tis & Extracted \\
\hline
HeiCo \cite{maier2021heidelberg} & 30 & 1499750 & Phase & Colon Rectum & - & - & - & - & Yes \\ 
Cholec80 \cite{twinanda2017endonet} & 80 & 176110 & Phase & Cholecyst & - & 7 & - & - & No \\ 
CholecT50 \cite{nwoye2022rendezvous} & 50 & 100863 & Triplet  & Cholecyst & 100 & 6 & 10 & 15 & No \\ 
MISAW \cite{arnaud2021micro} & 27 & 164277 & Phase, Step, Triplet  & Anastomose & 47 & 1 & 10 & 9 & No \\ 
Bypass40 \cite{sanat2021multi} & 40 & 26000 & Phase, Step & Stomach & - & - & - & - & No \\ 
PSI-AVA \cite{valderrama2022towards} & 8 & 73618 & Phase, Step, Triplet* & Prostate & 16 & 7 & 16 & - & No \\ 
ESAD \cite{vivek2020esad} & 4 & 33398 & Action & Cholecyst & - & - & 21 & - & No \\ 
RLLS12M \cite{zhao2022MURPHYRM} & 50 & 2113510 & Step, Task, Triplet  & Liver Cholecyst & 38 & 11 & 8 & 16 & No \\ 
Cataract \cite{lin2022instrument} & 20 & 13374 & Quintuple & Cataract & 32 & 12 & 15 & 12 & Yes \\ 
\hline
\textbf{PhacoQ (ours)} & 20 & 23580 & Quintuple & Cataract & 32 & 12 & 15 & 12 & No \\ 
\textbf{CholecQ (ours)} & 181 & 14480 & Quintuple & Cholecyst & 17 & 2 & 5 & 5 & No \\ 
\hline
\end{tabular}
\begin{tablenotes}
\footnotesize
\item[1] \# represents the number of, and Ins, Tis, Act, Avt is the abbreviation for instrument, tissue, action, activity.
\item[2] Triplet represents $\langle$instrument class, action class, tissue class$\rangle$, Triplet* represents $\langle$instrument class, instrument bounding box, action class$\rangle$, and Quintuple represents $\langle$instrument class, instrument bounding box, action class, tissue class, tissue bounding box$\rangle$. Extracted means that the video frames have been non-uniformly sampled.
\end{tablenotes}
\end{threeparttable}
\end{table*}

As summarized in Table \ref{tab_dataset}, existing surgical workflow datasets mainly focus on coarse-grained workflow representation and no dataset matches our task. To evaluate the efficiency of our algorithm, we build two datasets based on different surgical scenarios, including a private Cataract surgery and a publicly available Cholecystectomy surgery. We label the two datasets under the direction of doctors. Their detailed illustrations are listed in the following.
\subsection{Cataract Surgery Video (PhacoQ) Dataset}
We build a cataract surgery video dataset, named PhacoQ Dataset, based on phacoemulsification, which is a widely used cataract surgery. The dataset consists of 20 surgery videos with a frame rate of 1 fps. We make the dataset to be more suitable for actual surgical scenes by adding non-interaction frames. 
These videos are labeled frame by frame with five annotators directed by ophthalmologists. The annotation for each frame includes the location of instruments and tissues as well as the type of instruments, tissues, and actions. Twelve kinds of instruments, 12 kinds of tissues, and 15 kinds of actions form 32 kinds of interaction labels. 
PhacoQ Dataset is randomly split into a training set with 12 videos, a valid set with 4 videos, and a testing set with 4 videos. Some examples are presented in the first row of Fig. \ref{fig_result}.

\subsection{Cholecystectomy Surgery Video (CholecQ) Dataset}
We construct a cholecystectomy surgery video dataset, named CholecQ Dataset, based on two publicly available datasets CholecT50 \cite{nwoye2022rendezvous} and CholecSeg8k \cite{hong2020cholecseg8k}, both of which use the endoscopic images from the Cholec80 dataset \cite{twinanda2017endonet}. The data are obtained from laparoscopic cholecystectomy surgeries at the University Hospital of Strasbourg, France. The dataset is also labeled with five markers, including the bounding boxes and categories of instruments, the bounding boxes and categories of tissues, and the categories of actions. We have labeled two kinds of instruments, five kinds of tissues, and five kinds of actions, forming 17 kinds of interaction. The CholecQ Dataset contains 181 video snippets with 17 kinds of instrument-tissue interaction annotations and each video snippet consists of 80 image frames. The surgical video dataset is split into a training set with 145 video snips and a test set with 36 video snips. 
Some examples are presented in the first row of Fig. \ref{fig_result}.

\section{Proposed Method}
\begin{figure}
\centerline{\includegraphics[width=\linewidth]{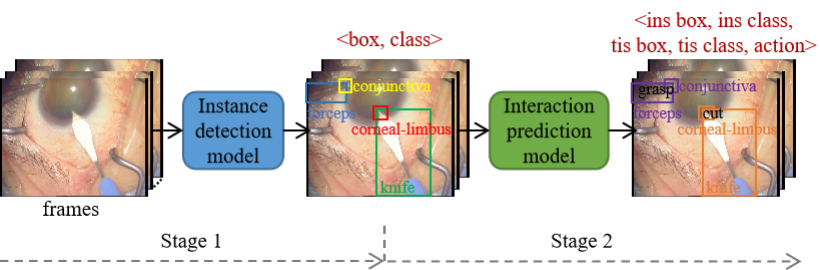}}
\caption{Pipeline of the proposed ITIDNet models. Instruments and tissues are detected in the first stage while actions for instrument-tissue pairs are predicted in the second stage.} 
\label{fig_overall} 
\end{figure}

The pipeline of the proposed Instrument-Tissue Interaction Detection Network (ITIDNet) is shown in Fig. \ref{fig_overall}. The framework follows a two-stage strategy with an instance detection stage and an interaction prediction stage. In the first stage, bounding boxes of instruments and tissues are detected using the instance detection model. In the second stage, these detected instruments and tissues are sent into the proposed interaction prediction model to predict interactions. 

\subsection{Instance Detection}

\begin{figure*}[!t]
\centerline{\includegraphics[width=\linewidth]{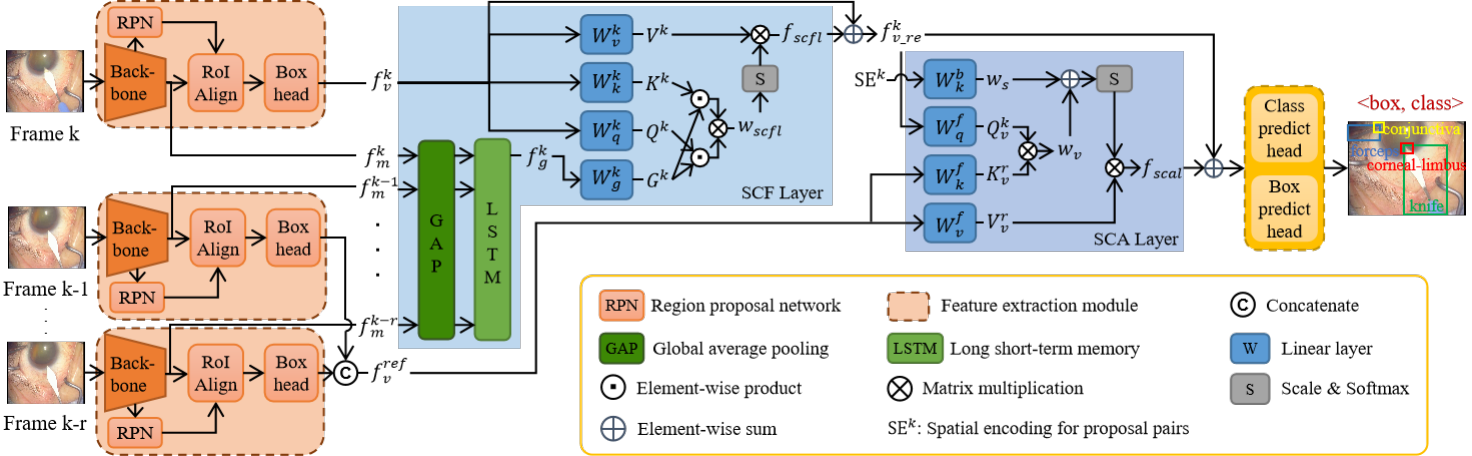}}
\caption{Illustration of the proposed instance detection model in the first stage. Extracting regional features using the Faster R-CNN backbone network, refining proposal features through the proposed SCF Layer and SCA Layer, and predicting the category and location of instruments and tissues. The proposed SCF Layer fuses global context information from the video snippet and exploits relationships of RoIs in the key frame $k$. The proposed SCA Layer utilizes relationships of RoIs in adjacent frames (from frame $k-r$ to $k$) guided by spatial encoding.} 
\label{fig_stage1} 
\end{figure*}
Due to poor image quality caused by movement or bleeding, mutual obscuration between instruments and tissues, and tissues with unclear boundaries, such as the transparent corneal in cataract surgeries, directly applying the off-the-shelf object detector to our surgery scenes cannot provide desired detection results. 
Under these circumstances, relying solely on the visual proposal features itself is not sufficient to detect the object well. Combining global context information in the video snippet with visual features is a good way to improve performance. 
Furthermore, as medical prior knowledge, instruments and their corresponding tissues in the surgical process must meet certain standards. Therefore, we propose to aggregate features among instruments and tissues in the same frame and between adjacent frames to improve detection accuracy.

As shown in Fig. \ref{fig_stage1}, 
we take the current frame $k$ as the key frame and frame $k-r$ to $k-1$ as reference frames $ref$. A video snippet consists of the key frame and reference frames, with $r+1$ frames in total. 
The instance detection model follows the basic structure of Faster R-CNN, using a backbone network to extract feature maps $f_m\in\mathbb{R}^{c\times h\times w}$ for each input frame first, where $c$, $h$, and $w$ denotes the number of channels, height, and width for the feature maps. A Region Proposal Network (RPN) is used to generate a set of region proposals based on these feature maps. 
Then an RoI align network followed by linear box heads extract regional features as visual features $f_v\in\mathbb{R}^{N_p\times c}$, where $N_p$ is the number of proposals. 
A Snippet Consecutive Feature (SCF) Layer and a novel Spatial Corresponding Attention (SCA) Layer are proposed to exploit instrument-tissue relations for feature refinement.
For each proposal, a class prediction head predicts its probabilities belonging to a certain instrument or tissue category and a box prediction head refines the bounding box of this proposal via regression.

\subsubsection{Snippet Consecutive Feature (SCF) Layer}
In the surgery scenes, poor picture quality, caused by bleeding, reflections, or mutual obscuration of instruments and tissues, hinders accurate object detection. Since the context information of a video snippet provides the overall global information of the video snippet, it is considered as a good aid to improve detection performance when object appearance features alone are not sufficient for object detection. 
If the model can capture global context information about the video snippet, it will be easier to detect objects since the information about a short video snippet will imply the presence of instruments and tissues. For example, since the artificial lens is transparent, it is difficult to detect properly. However, a video snippet of artificial lens implantation will imply that there will be an artificial lens in the image. 
Besides, in a standard surgical procedure, there is a correspondence between the instrument and the tissue that it interacts with. For instance, the presence of a knife implies the presence of forceps and the target of the knife is usually corneal-limbus in phacoemulsification. These correspondences can be used to assist in detecting instruments and tissues through feature refinement. 
Thus, we design a Snippet Consecutive Feature (SCF) Layer to combine snippet context information and exploit instrument-tissue relationships of intra-frame for feature refinement. 

As shown in Fig. \ref{fig_stage1}, feature maps of frames in the video snippet $f_m^{k-r}$, ..., $f_m^{k-1}$, $f_m^k\in\mathbb{R}^{c\times h\times w}$ are sent to the Global Average Pooling Layer and Long Short-Term Memory (LSTM) Layer to obtain the snippet context features $f_g^k\in\mathbb{R}^{1\times c}$. These obtained snippet context features include temporal information about the key frame and reference frames.

To combine global context features of the video snippet with instance features and model the relationships between instances in the current frame, the snippet context features $f_g^k$ are mapped into $G^k\in\mathbb{R}^{1\times c}$ via linear mapping layers $W_g^k$. Also, visual features of proposals in the key frame $f_v^k$ are mapped into a query $Q^k\in\mathbb{R}^{N_p\times c}$, a key $K^k\in\mathbb{R}^{N_p\times c}$, and a value $V^k\in\mathbb{R}^{N_p\times c}$ via linear mapping layers $W_q^k$, $W_k^k$, and $W_v^k$. The attention scores $w_{scfl}$ can be calculated by: 
\begin{equation}
    w_{scfl} = \frac{{(Q^k\cdot{G^k})}\,{{(K^k\cdot{G^k})}^T}}{\sqrt{d^k}}
\end{equation}
\noindent where $T$ is transpose function and $d^k$ is the dimension of $K^k$. 

The output attention features of SCF Layer $f_{scfl}$ are computed by the multiplication of the mapped RoI visual features $V^k$ and the attention scores $w_{scfl}$, formulated as
$f_{scfl} = \mathrm{softmax}(w_{scfl}) V^k$.

\subsubsection{Spatial Corresponding Attention (SCA) Layer} 
In surgical scenes, it is common for instruments and tissues to be partially invisible due to mutual occlusion and the movement of instruments. In this case, detecting instruments and tissues using proposal features alone cannot produce accurate detection results. Since in such a short video snippet the instruments or tissues may reveal different parts in different frames, we suppose that information from adjacent frames can be helpful for improving the detection performance based on feature aggregation. 
In order to obtain feature aggregation weights, spatial location information is also vital in addition to visual features. Typically, the instrument and its corresponding tissue are relatively close. 
Thus, to incorporate the information of proposals in the adjacent frames for feature aggregation, we design a Spatial Corresponding Attention (SCA) Layer.

As illustrated in Fig. \ref{fig_stage1}, spatial location information for proposal pairs between the key frame and the reference frames are encoded as the spatial encoding. The spatial encoding is the concatenation of normalized center coordinates of the proposal boxes, normalized proposal box width and height, normalized proposal box area, normalized proposal box aspect ratio, and pairwise relationships. The pairwise relationships include the intersection over union and the differences between the center coordinates of the proposal boxes normalized by the dimensions of the proposal box in the key frame. 
The inputs to SCA Layer are the refined visual features of the proposals in the key frame $f_{v\_re}^k\in\mathbb{R}^{N_p\times c}$, the concatenated visual features of the proposals in the reference frames $f_v^{ref}\in\mathbb{R}^{r*N_p\times c}$, and the spatial encoding for proposal pairs $SE^k\in\mathbb{R}^{N_p\times N_p\times 16}$.

Specifically, the spatial encoding of proposal pairs $SE^k$ is sent into a fully-connected network $W^b_k$ to compute a spatial weight $w_s = \mathrm{FC}(SE^k) $ of proposals from the key frame and reference frames. 
To obtain the spatial correspondence, visual features of proposals in the key frame $f_{v\_re}^k$ are mapped into a query $Q_v^k\in\mathbb{R}^{N_p\times c}$ via linear layers $W_q^f$. Concatenated visual features of proposals in $r$ reference frames $f_v^{ref}$ are also mapped into a key $K_v^r\in\mathbb{R}^{r*N_p\times c}$ and a value $V_v^r\in\mathbb{R}^{r*N_p\times c}$ via linear layers $W_k^f$ and $W_v^f$, as shown in Fig. \ref{fig_stage1}. The visual weight $w_v$ of proposals in the key frame and its reference frames is calculated by $w_v = \frac{{Q_v^k}\,{{K_v^r}^T}}{\sqrt{d_v^r}}$, where $T$ is the transpose function, and $d_v^r$ denotes the dimension of $K_v^r$. 

Finally, the attention weight for feature aggregation in SCA Layer is the sum of the spatial weight $w_s$ and the visual weight $w_v$. The output features of SCA Layer $f_{scal}$ are computed by the multiplication of the mapped RoI features of the reference frames $V_v^r$ and the attention weights, formulated as 
$f_{scal} = \mathrm{softmax}(w_v + w_s)V_v^r$.

\subsection{Interaction Prediction}
\begin{figure*}[!t]
\centerline{\includegraphics[width=\linewidth]{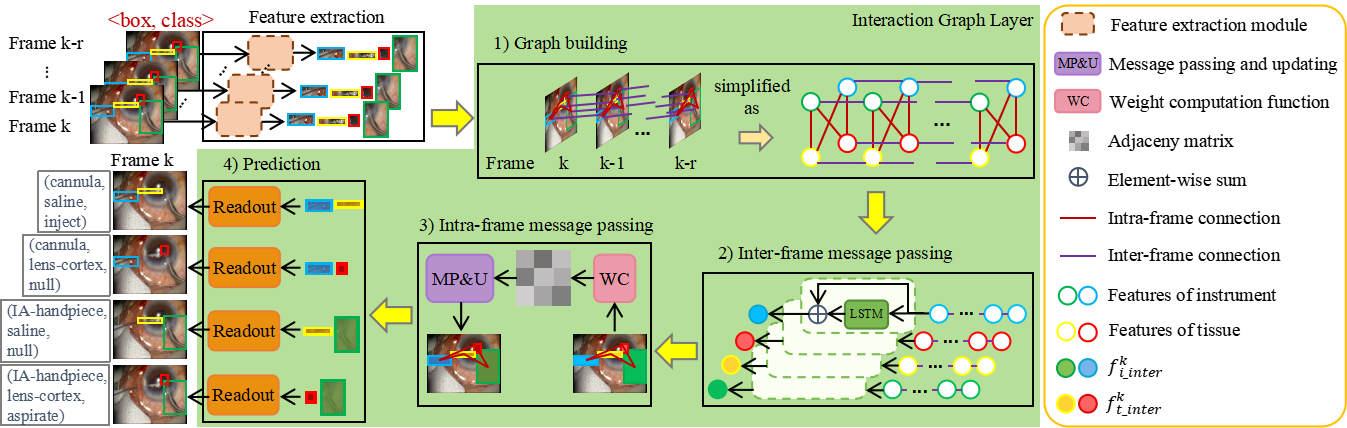}}
\caption{Illustration of the proposed Interaction Prediction model in the second stage. Instruments and tissues detected in stage 1 are sent to the feature extraction module to extract regional visual features first. These features are sent to the Temporal Graph Layer to predict actions between instruments and tissues. In this layer, an interaction graph is built to represent the instruments and tissues and the relationships between them. 
} 
\label{fig_stage2} 
\end{figure*}

The model framework of the interaction prediction stage is shown in Fig. \ref{fig_stage2}. A feature extraction module is utilized to extract visual features of detected instruments and tissues and a Temporal Graph (TG) Layer is proposed to predict the actions between the detected instruments and tissues by modeling inter-frame and intra-frame relationships of them. 

To extract visual features of instruments and tissues, we directly adopt the backbone network in stage 1 to extract feature maps for video frames $k-r$ to $k$. 
The detected bounding boxes combined with the extracted feature maps are passed through RoI Align to extract regional features. These regional features are then sent to a Box Head which includes a flatten operation and a fully-connected layer with ReLU activation function to extract instance features $f^n_o\in\mathbb{R}^{N\times c}$, where $n=\{ref,k\}$ represents reference (from frame $k-r$ to $k-1$) and key frames ($k$), $o=\{i, t\}$ represents the detected object (instrument or tissue). $N=\{N_i, N_t\}$, $N_i$ is the number of detected instruments $i$, and $N_t$ is the number of detected tissues $t$ in frame $n$. 

As relation reasoning is vital for interaction prediction, we propose a graph-based model, Temporal Graph Layer, to build multiple relationships of instruments and tissues. Firstly, we build an interaction graph to connect detected instruments and tissues in a video snippet. Then we refine graph features by exploring relationships between graph nodes, including the intra-frame relation between instruments and tissues, and the inter-frame relation of the same instruments or tissues. Finally, actions between instruments and tissues are predicted based on the refined interaction graph.

\subsubsection{Graph building}
An interaction graph is constructed for a video snippet to represent the instruments and tissues and the relationships between them. The graph is constructed by nodes, intra-frame connection, and inter-frame connection. \textbf{Nodes} are the instruments and tissues detected in the first stage and their initial features are the input regional visual features. The \textbf{intra-frame connection} is bidirectional and defined as the correlation between instruments and tissues in one frame. 

 In surgical scenes, there are often multiple objects of the same category in one frame. In order to predict their respective actions, we propose to analyze them one by one. We assume every object (tissue or instrument) in one frame as an instance, $i=\{i_1, i_2, .., i_{N_i}\}$ denotes the set of detected instrument instance, $t=\{t_1, t_2, .., t_{N_t}\}$ denotes the set of detected tissue instance. As instruments or tissues will continue to appear in the surgical scene for a period of time, the instance is considered to be continuous in a short snippet. We define the \textbf{inter-frame connection} as the temporal correlation of the same instance between the keyframe and reference frames. 
 
\begin{figure}[!h]
\centerline{\includegraphics[width=\columnwidth]{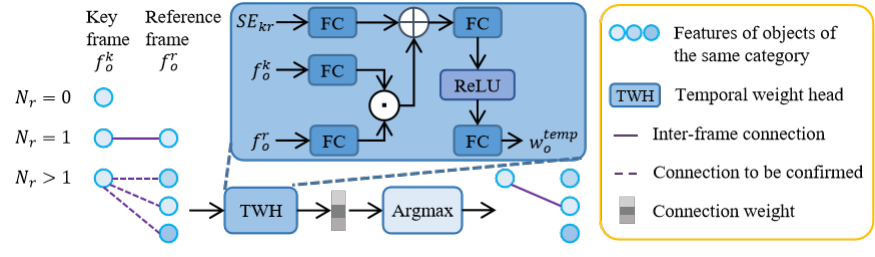}}
\caption{Illustration of making the inter-frame connection for the instance $o$ between the key frame and one reference frame $r$. $N_r$ denotes the number of objects, which have the same class as $o$, in frame $r$.} 
\label{fig_intra} 
\end{figure}

For every object in the set $o=\{i_1, i_2, .., i_{N_i},t_1, t_2, .., t_{N_t}\}$ of the current key frame, the first step of the algorithm is to find the same type of object in each reference frame.
As shown in Fig. \ref{fig_intra}, we define the number of objects with the same category of $o$ in a reference frame as $N_r \le {N_i} \mathrm{or} {N_t}$. If $N_r$ equals zero, there is no connection between $o$ and other objects in the reference frame. If $N_r$ equals one, the object in the reference frame is regarded as the same instance as $o$, and there is an inter-frame connection between $o$ and the same instance in the reference frame. If $N_r$ is larger than one, the connection among these objects should be determined. In other words, we need to determine which object in the reference frame is the same instance as $o$ in the keyframe. 
Therefore, we propose a Temporal Weight (TW) Head to generate the inter-frame connection weight $w^{temp}_o$, defined as: 
$w^{temp}_o = \mathrm{FC}(\mathrm{ReLU}(\mathrm{FC}(\mathrm{FC}(f^k_o) * \mathrm{FC}(f^r_o) + \mathrm{FC}(SE_{kr}))))$
, $f^k_o\in\mathbb{R}^{1\times c}$ is the regional visual features of an instance $o$ in the keyframe, $f^r_o\in\mathbb{R}^{N_r\times c}$ denotes regional visual features of objects in the reference frame which have the same category as the instance $o$, and $SE_{kr}\in\mathbb{R}^{1\times N_r\times 16}$ is the spatial encoding of the object pairs. 
The index of the same instance in the reference frame is calculated by $argmax$ of the connection weight.

\subsubsection{Inter-frame message passing}
An action verb is a high-level abstraction of the interaction between instruments and tissues over a period of time. The temporal information, such as the motion of the same instance in consecutive frames, is of great significance for action prediction. Hence, in the interaction graph, we pass and update messages between inter-frame connections to exploit temporal information of action. 

After sorting out inter-frame connections in the graph, visual features of the same instances in the key frame $f_o^k$ and reference frames $f_o^{k-r}, ..., f_o^{k-1}$ are sent to LSTM to capture temporal information for feature refinement. We adopt LSTM as the message passing function and the residual connection as the update function to obtain the refined node features in the key frame $f^k_{o\_inter}\in\mathbb{R}^{1\times c}$ as shown in Fig. \ref{fig_stage2}.

\subsubsection{Intra-frame  message passing}
Since instruments and tissues in the same frame should be related, we model their relationships through intra-frame connections.  
Intra-frame graph weights $w^k_{i,t}$ that indicates the relationships between instruments and tissues in the key frame can be calculated as:
\begin{equation}
 w^k_{i,t} = \mathrm{FC}(\mathrm{ReLU}(\mathrm{FC}(\mathrm{FC}(f^k_i) * \mathrm{FC}(f^k_t) + \mathrm{FC}(SE_{it}))))
\end{equation}

\noindent where $f^k_i\in\mathbb{R}^{N_i\times c}$ and $f^k_t\in\mathbb{R}^{N_t\times c}$ denotes node features of instruments and tissues in the key frame, and $SE_{it}\in\mathbb{R}^{N_i\times N_t\times c}$ denotes the spatial encoding of instrument-tissue pairs. Noted that the inputs to intra-frame message passing are the outputs after inter-frame message passing. To simplify the expression, the inputs are not written with subscripts $\_{inter}$.

To generate more discriminative features, we propose to refine node features of instruments and tissues in the key frame by bidirectional message passing and updating functions. The node features of instruments in the key frame are updated as: 
$f^k_{i\_intra} = \mathrm{Norm}(f^k_i + \sum_{t=1}^{N_t} w_{i,t}f^k_t)$, 
$f^k_{i\_intra}$ denotes output features of instrument's node and $\mathrm{Norm}(\cdot)$ is the LayerNorm operation.

The nodes of tissues in the key frame are updated as: 
$f^k_{t\_intra} = \mathrm{Norm}(f^k_t + \sum_{i=1}^{N_i} w_{t,i}f^k_i)$, 
$w_{t,i}$ is the transpose of $w_{i,t}$ and $f^k_{t\_intra}$ denotes output features of tissue's node. 

\subsubsection{Action prediction}
After message passing and updating between inter-frame connections and intra-frame connections, the action of the instrument-tissue pair is calculated using the readout function. In addition to instrument and tissue node features, spatial encoding of the instrument-tissue pair and global feature map $f^k_m$ of the key frame are helpful in predicting the action $s_a$. The calculation process for $s_a$ is formulated as:
\begin{equation}
s_a = \mathrm{Sigmoid}(\mathrm{FC}(\mathrm{readout}(f^k_{i\_intra}, f^k_{t\_intra}, SE_{it}, f^k_m))).
\end{equation}
\noindent where readout(·) is defined as $\mathrm{ReLU}(\mathrm{Norm}(\mathrm{FC}(\mathrm{FC}(f^k_{i\_intra}) \\ + \mathrm{FC}(f^k_{t\_intra}) + \mathrm{FC}(SE_{it}) + \mathrm{FC}(\mathrm{GAP}(f^k_m)))))$ and GAP(·) is global average pooling.

Since the surgery is often standardized, the action types for certain types of instruments and tissues are limited. For post-processing, since the categories of instruments and tissues are already decided in the first stage, a prior score can be used for more accurate prediction. A certain type of instrument or tissue can be mapped into several types of possible actions. The score of the possible type of mapping action is set to the same as the score of the instrument or tissue, denoted as $s_i$ and $s_t$. The final interaction score $s$ is calculated as $s = s_a*s_i*s_t.$ 
In summary, the quintuple outputs for instrument-tissue interaction detection are obtained by pairing the instruments and tissues detected in the first stage and generating the action predictions between them in the second stage. 

\subsection{Optimization}
The instance detection model and the interaction prediction model are trained separately.
To train the instance detection model, we just adopt the same loss as Faster R-CNN \cite{ren2015faster}, including the RPN loss and the Fast R-CNN loss. 
For model training in the interaction prediction model, we first assign each prediction interaction with ground-truth labels. Specifically, Intersection over Union (IoU) is calculated between predicted and ground-truth instruments (denoted as $\mathrm{IoU_i}$) and between predicted and ground-truth tissues (denoted as $\mathrm{IoU_t}$). If the minimum value between $\mathrm{IoU_i}$ and $\mathrm{IoU_t}$ for the interaction prediction is less than a threshold, then the action label is assigned as the ground-truth action label. To alleviate the class imbalance problem, the focal loss $\mathcal{L}_{focal}$ \cite{lin2020focal} is applied for model training.

\section{Experiments and Results}
\subsection{Experimental Settings}
\subsubsection{Evaluation Metrics}
Mean Average Precision (mAP) \cite{gupta2015visual}, which is widely used in evaluating the human-object interaction detection task, is adopted to evaluate the proposed network for instrument-tissue interaction detection. A detected interaction is considered as a true positive detection when the predicted instrument-tissue-action class is correct and both the predicted instrument and tissue boxes have an Intersection over Union (IoU) higher than 0.5 with the corresponding ground truth pair.
To validate the effectiveness of the model in the first stage, mAP is also used to measure instrument and tissue detection performance in the first stage. We denote mAP for the instrument and tissue detection as $\mathrm{mAP_{IT}}$ and mAP for the instrument-tissue interaction detection as $\mathrm{mAP_{ITI}}$.

\subsubsection{Implementation Details}
Input video frames for all models maintain the original aspect ratio and are reshaped to a height of 448. Considering the balance of effectiveness and efficiency, the number of reference frames is set to 3. Experiments on the number of reference frames are presented in section \ref{F}. ResNet50-FPN is adopted as the backbone network in our framework. In the first stage, our models are trained for 20 epochs and the initial learning rate is set to 0.001 with 0.1 decayed at the 10th and 15th epochs. Detection results of the first stage with scores lower than 0.2 are filtered out and non-maximum suppression with a threshold of 0.5 is applied. Up to 5 instrument boxes and 5 tissue boxes with the highest score are selected for the second stage of interaction prediction. In the second stage, our models are trained for 20 epochs with the initial learning rate at 0.0001. SGD optimizer with 0.9 momentum and 0.0001 weight decay is used in both stages. Our models are trained on GeForce GTX 2080 Ti GPUs.

\subsection{Comparison Experiments}
We compare the proposed framework with several state-of-the-art methods based on the PhacoQ Dataset and the CholecQ Dataset. Inspired by \cite{lin2022instrument}, we design a Baseline model that uses Faster R-CNN \cite{ren2015faster} for instrument and tissue detection and a fully-connected layer to predict actions for instrument-tissue pairs. Two state-of-the-art human-object interaction detection methods, iCAN \cite{gao2018ican} and Zhang et al. \cite{zhang2021spatially}, are reimplemented for instrument-tissue interaction detection. The QDNet \cite{lin2022instrument} is the first method for instrument-tissue interaction detection. Quantitative results on the PhacoQ test set and the CholecQ test set are shown in Table \ref{tab_compare}. As Faster R-CNN is used as the first stage model in both iCAN \cite{gao2018ican} and Zhang et al. \cite{zhang2021spatially}, they have the same $\mathrm{mAP_{IT}}$ for the first stage.

\begin{table}\centering
\caption{Comparison results on PhacoQ and CholecQ Dataset.} 
\label{tab_compare}
\begin{tabular}{|p{60pt}|p{28pt}|p{28pt}|p{28pt}|p{28pt}|}
\hline
\multirow{2}{*}{Methods} & \multicolumn{2}{c|}{PhacoQ Dataset} & \multicolumn{2}{c|}{CholecQ Dataset} \\
\cline{2-5}
 & $\mathrm{mAP_{IT}}$ & $\mathrm{mAP_{ITI}}$ & $\mathrm{mAP_{IT}}$ & $\mathrm{mAP_{ITI}}$\\
\hline
Baseline \cite{ren2015faster} & 58.17\% & 33.75\% & 59.72\% & 28.69\% \\
iCAN \cite{gao2018ican} & 58.17\% & 34.13\% & 59.72\% & 31.73\%\\
Zhang et al. \cite{zhang2021spatially} & 58.17\% & 34.63\% & 59.72\% & 35.27\%\\
QDNet \cite{lin2022instrument} & 58.66\% & 34.89\% & 61.79\% & 36.83\% \\
\hline
\textbf{ITIDNet (Ours)} & \textbf{60.21\%} & \textbf{36.82\%} & \textbf{63.36\%} & \textbf{39.01\%} \\
\hline
\end{tabular}
\end{table}

The proposed method outperforms other methods by a large margin in both instance detection and interaction prediction stages. With SCF Layer and SCA Layer, the proposed method boosts the $\mathrm{mAP_{IT}}$ for the first stage by 2.04\% on PhacoQ Dataset and 3.64\% on CholecQ Dataset. Although QDNet also emphasizes the importance of spatial and temporal relationships between proposals in adjacent frames, our proposed method achieves better results by combining snippet context information to exploit the relationships between proposals in the key frame. Furthermore, thanks to the relation reasoning in the second stage, the improvement for the overall instrument-tissue interaction detection task is more significant as 3.07\% higher $\mathrm{mAP_{ITI}}$ on PhacoQ Dataset and 10.32\% $\mathrm{mAP_{ITI}}$ on CholecQ Dataset compared with the baseline model. By wisely incorporating the inter-frame connection in the graph structure, our proposed framework performs better than the QDNet, which only takes the intra-frame connection into consideration. Overall, the proposed method has advantages in detecting instrument-tissue interaction, especially in action prediction.

\begin{figure*}[!t]
\centerline{\includegraphics[width=\linewidth]{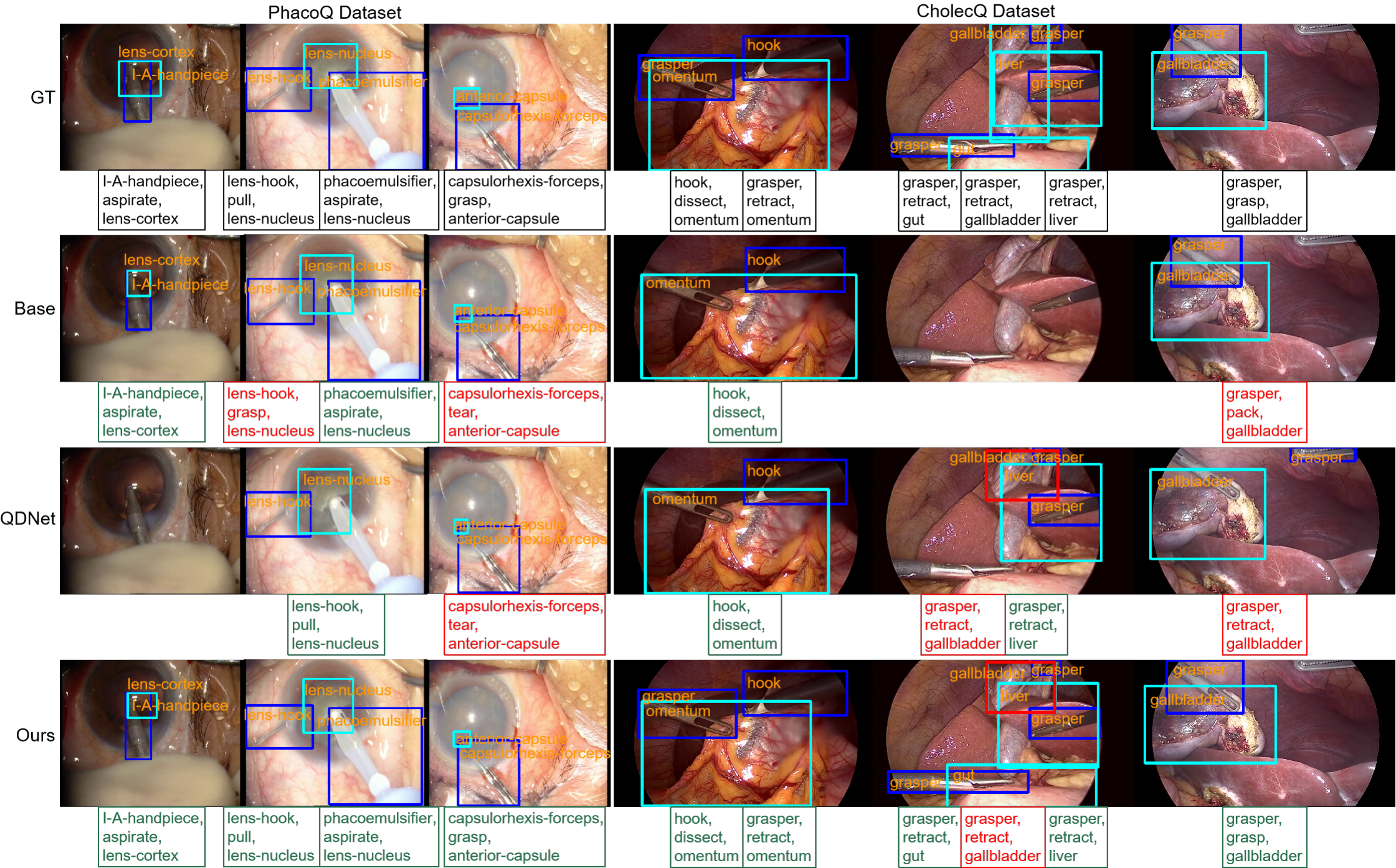}}
\caption{Example results on PhacoQ Dataset and CholecQ Dataset. From top to bottom, we represent the ground-truth, results of the Faster R-CNN baseline model \cite{ren2015faster}, QDNet \cite{lin2022instrument}, and our model respectively. The ground-truth or detection bounding boxes of instruments and tissues are marked in blue and light blue. False detection bounding boxes are marked in red. Incorrect interaction detection results are marked in red while correct detection results are marked in green. Only instrument-tissue interactions with an interaction score $s$ greater than 0.5 are presented.}
\label{fig_result}
\end{figure*}

Qualitative results of the Baseline, QDNet, and our model are presented in Fig. \ref{fig_result}. The results prove that the proposed model can provide more accurate detection of instrument-tissue interactions than other methods. Especially, when both the instrument and tissue are correctly detected, the proposed model is able to more accurately determine the action between them. Note that in the fifth column of Fig. \ref{fig_result}, QDNet and our proposed model only detect half of the gallbladder tissue due to the obscuration of the liver. The category labels for this interaction are correct, however, the bounding boxes of the tissue detection results and the gound-truth are different.

\subsection{Ablation Studies on Key Components of Framework}
Ablation studies on PhacoQ Dataset and CholecQ Dataset are conducted to validate the effectiveness of the three key components. The results are listed in Table \ref{tab_block}. Our framework achieves the best performance with 60.21\% $\mathrm{mAP_{IT}}$ and 36.82\% $\mathrm{mAP_{ITI}}$ on PhacoQ Dataset. Removing the SCF Layer decreases the performance by 0.5\% at $\mathrm{mAP_{IT}}$ and further leads to the reduction of $\mathrm{mAP_{ITI}}$ by 0.94\%. Removing the SCA Layer from the proposed framework impairs the performance even more by 1.16\% at $\mathrm{mAP_{IT}}$ and 1.95\% at $\mathrm{mAP_{ITI}}$. Besides, replacing the TG Layer by simply concatenating the features of the instrument and tissue pairs results in a 1.89\% drop at $\mathrm{mAP_{ITI}}$. Similar trends were obtained on the CholecQ dataset for removing the SCF Layer and SCA Layer while removing TG Layer has a greater impact (a 7.52\% $\mathrm{mAP_{ITI}}$ drop) on the CholecQ dataset. This may be due to the fact that for an instrument, there are more interaction combinations on the CholecQ dataset. Hence, the TG Layer becomes even more important as it is responsible for interaction prediction. Therefore, our proposed layers: SCF Layer, SCA Layer, and TG Layer are proved to be useful in improving the accuracy of instrument-tissue interaction detection.

\begin{table}
\centering
\caption{Ablation study on key modules in the proposed framework.}
\label{tab_block}
\begin{tabular}{c c c|c|c|c|c}
\hline
\multicolumn{3}{c|}{Layers} & \multicolumn{2}{c|}{PhacoQ Dataset} & \multicolumn{2}{c}{CholecQ Dataset}  \\
\hline
 SCF & SCA & TG & $\mathrm{mAP_{IT}}$ & $\mathrm{mAP_{ITI}}$ & $\mathrm{mAP_{IT}}$ & $\mathrm{mAP_{ITI}}$\\
\hline
\checkmark & \checkmark & \checkmark & \textbf{60.21\%} & \textbf{36.82\%} & \textbf{63.36\%} & \textbf{39.01\%} \\
 & \checkmark & \checkmark & 59.71\% & 35.88\% & 61.34\% & 38.77\% \\
\checkmark & & \checkmark & 59.05\% & 34.87\% & 60.41\% & 35.24\% \\
\checkmark & \checkmark & & 60.21\% & 34.93\% & 63.36\% & 31.49\% \\
\hline
\end{tabular}
\end{table}

\subsection{Experiments for Designing the First Stage Model}
\subsubsection{SCF Layer}
We conduct experiments on the structure of the SCF Layer to verify the effectiveness of snippet context information and the relationships of RoIs in the key frame for instrument and tissue detection, as shown in the first part of Table \ref{tab_1}. A \textbf{GC Layer} is designed to utilize the global context features of each frame by simply concatenating it with the RoI features. On the basis of the Faster R-CNN baseline model, GC Layer improves the performance of the model at detecting instruments and tissues by 0.17\% $\mathrm{mAP_{IT}}$. This proves the effectiveness of global context information. To validate the effectiveness of the relationships of RoIs in the key frame for instrument and tissue detection, on top of the baseline model, we add an \textbf{Attention Layer}, which is a modified version of SCF Layer with the snippet context features removed. It can be formulated as $f_{al} = \mathrm{softmax}(\frac{{Q^k}\,{{K^k}^T}}{\sqrt{d}}) V^k$. The $\mathrm{mAP_{IT}}$ improves by 0.63\%, rising to 58.80\%. 
Adding the proposed SCF Layer on the baseline achieves the best performance with 59.05\% $\mathrm{mAP_{IT}}$, which proves the effectiveness of the way we use global context information to help reason the relationships of RoIs in the key frame. Moreover, to prove the need of using global context features of the video snippet, we build a new \textbf{SCF Layer without LSTM}, which utilizes global context features from the key frame, to replace SCF Layer. As a result, the performance of the model drops to 58.56\%.

\begin{table}
\centering
\caption{Experiments for the design of SCF Layer and SCA Layer}
\label{tab_1}
\begin{tabular}{p{55pt}|p{100pt}|p{25pt}}
\hline
\multicolumn{2}{c|}{Design} & $\mathrm{mAP_{IT}}$ \\
\hline
\multicolumn{2}{l|}{Baseline (Faster R-CNN)} & 58.17\%\\
\hline
 & + GC Layer &  58.34\%\\
Experiments for & + Attention Layer & 58.80\% \\
SCF Layer & + SCF Layer without LSTM &  58.56\%\\
& + SCF Layer &  59.05\%\\
\hline
& + CA Layer &  59.44\%\\
Experiments for & + SCA Layer & 59.67\%\\
SCA Layer & + SCF Layer + CA Layer & 59.73\% \\
& + SCF Layer + SCA Layer & \textbf{\textbf{60.21\%}} \\
\hline
\end{tabular}
\end{table}

\subsubsection{SCA Layer}
In the second part of Table \ref{tab_1}, we conduct experiments on the structure of the proposed SCA Layer to validate the effectiveness of spatial information and the relationships of RoIs in adjacent frames. We build a \textbf{CA Layer}, which is formulated as $f_{cal} = \mathrm{softmax}(\frac{{Q_v^k}\,{{K_v^r}^T}}{\sqrt{d_v^r}}) V_v^r$. With CA Layer, the $\mathrm{mAP_{IT}}$ increases to 59.44\% compared with the Faster R-CNN baseline model. This proves that the relationships of RoIs in adjacent frames are effective in helping improve detection. The proposed SCA Layer can be viewed as CA Layer guided by spatial information. Compared with the non-spatial information version, the baseline model with SCA Layer further enhances the detection performance to 59.67\%. This indicates that spatial information is also helpful in detection. Moreover, our model is able to perform even better with the addition of SCF Layer.

\subsection{Experiments for Designing the Second Stage Model}
As shown in Table \ref{tab_graph}, we conduct ablation studies on intra-frame connection and inter-frame connection in TG Layer: (1) \textbf{Base}: we use the output instruments and tissues in the first stage as the graph nodes and only the readout function is used in the second stage to predict the actions for the instrument-tissue pairs; (2) \textbf{Intra}: we employ temporal graph network with only intra-frame connection in the proposed model; (3) \textbf{Intra'}: we replace the weight calculation function in Intra with $w_{i,t}^k = \mathrm{FC}(\mathrm{relu}(\mathrm{FC}(\mathrm{Concat}(f_i, f_t))))$; (4) \textbf{Inter}: we employ temporal graph network with only inter-frame connection in the proposed model; (5) \textbf{Inter'}: we replace the TW Head in Inter with the linear layer which can be formulated as $w = \mathrm{FC}(\mathrm{relu}(\mathrm{FC}(\mathrm{Concat}(f_k, f_r))))$. (6) Ours: we use both Intra and Inter as described in the methodology section. 

\begin{table}[h]
\centering
\caption{Ablation Study on the graph structure}
\label{tab_graph}
\setlength{\tabcolsep}{3pt}
\begin{tabular}{c|c}
\hline
Graph structure & $\mathrm{mAP_{ITI}}$ \\
\hline
Base & 34.93\% \\
Intra' & 35.21\% \\
Intra & 35.47\% \\
Inter' & 36.01\% \\
Inter & 36.16\% \\
\hline
\textbf{Ours} (Intra+Inter) & \textbf{36.82\%} \\
\hline
\end{tabular}
\end{table}
It is observed that the Base model obtains reasonable detection results with 34.93\% $\mathrm{mAP_{ITI}}$. Adding the intra-frame connection without spatial encoding weight calculation (Intra') boosts the performance with 0.28\% gain in $\mathrm{mAP_{ITI}}$. Furthermore, with the assistance of spatial encoding in weight calculation (Intra), the $\mathrm{mAP_{ITI}}$ rises to 35.47\% with 0.54\% gain. Besides, adding the inter-frame connection without spatial encoding in TW Head on the base model (Inter') improves the $\mathrm{mAP_{ITI}}$ by 1.08\% and a 1.23\% improvement occurred in $\mathrm{mAP_{ITI}}$ with spatial encoding in TW Head. With both inter-frame and intra-frame connections (ours), the proposed model achieves the highest performance with 36.82\% $\mathrm{mAP_{ITI}}$. These experimental results demonstrate the effectiveness of the inter-frame connection and intra-frame connection.
\begin{table}[h]
\centering
\caption{Ablation study on the number of reference frames}
\label{tab_seq}
\setlength{\tabcolsep}{8pt}
\begin{tabular}{c| c c c c}
\hline
$r$ & 1 & 3 & 5 & 7 \\
\hline
$\mathrm{mAP_{IT}}$ & 58.12\% & 60.21\% & 60.41\% & 60.05\% \\
$\mathrm{mAP_{ITI}}$ & 35.86\% & 36.82\% & 36.90\% & 36.41\% \\
\hline
$t_1$ (s) & 0.077 & 0.143 & 0.201 & 0.267 \\
$t_2$ (s) & 0.051 & 0.086 & 0.121 & 0.158 \\
\hline
\end{tabular}
\end{table}
\subsection{Experiments on the Number of Reference Frames}
\label{F}

Reference frames are core aids in improving model performance. In order to more comprehensively analyze the contribution of temporal information, we conduct experiments on the number of reference frames as shown in Table \ref{tab_seq}. As the number of reference frames grows, both $\mathrm{mAP_{IT}}$ and $\mathrm{mAP_{ITI}}$ of the proposed model gradually rise. It's also observed that the rate of growth in performance is decreasing. The inference time for one key frame for the instance detection stage ($t_1$) and the interaction prediction stage ($t_2$) are also provided. Considering the balance of effectiveness and efficiency, 3 is a wise choice for the number of reference frames.

\subsection{Statistical Significance Analysis}
The statistical significance of the performance of the proposed model is measured by comparing it with the baseline model and the state-of-the-art model QDNet. Besides, the statistical significance of key components of our ITIDNet is measured. The videos within the PhacoQ test set are segmented into clips, with each clip comprising 300 frames except for the last clip of a video. We calculated the $\mathrm{mAP_{ITI}}$ on each video clip in the PhacoQ test set and the CholecQ test set. The null hypothesis states that both algorithms perform equally well. Based on the obtained $\mathrm{mAP_{ITI}}$ score on each video clip, we used the Wilcoxon signed-rank test \cite{demvsar2006statistical} for significance analysis and calculated the p-value as presented in Table \ref{tab_p}. As the p-values for ITIDNet against the state-of-the-art methods on the PhacoQ test set and CholecQ test set are significantly smaller than 0.05, the null hypothesis can be rejected at a confidence level of 5\%. Therefore, there is a significant difference between our proposed models ITIDNet and other methods on both datasets. Besides, it can be seen from the table that our proposed key components are helpful for our quintuple detection.

\begin{table}[ht]
\centering
\caption{P-value in Wilcoxon signed-rank test of comparisons and key components}
\label{tab_p}
\setlength{\tabcolsep}{8pt}
\begin{tabular}{c| c c}
\hline
Method & PhacoQ dataset & CholecQ dataset \\
\hline
ITIDNet vs. baseline & 0.0052 & 0.0011 \\
ITIDNet vs. QDNet & 0.0494 & 0.0349 \\
ITIDNet vs. w/o SCA Layer & 0.0708 & 0.0946 \\
ITIDNet vs. w/o SCF Layer & 0.1323 & 0.0679 \\
ITIDNet vs. w/o TG Layer & 7.6e-06 & 0.0393 \\
\hline
\end{tabular}
\end{table}

\section{Conclusion}
In this paper, we propose to represent instrument-tissue interaction as $\langle$instrument class, instrument bounding box, tissue class, tissue bounding box, action class$\rangle$ quintuple for more comprehensive surgical scene understanding. To perform the instrument-tissue interaction detection task, a novel model named ITIDNet is proposed. The performance of the model is improved by (i) incorporating global context information of the video snippet with the relationships of RoIs in the key frame, (ii) adopting the relationships of RoIs between the key frame and reference frames guided by spatial encoding to assist in detecting instruments and tissues, and (iii) introducing a graph neural network with inter-frame connections and intra-frame connections to reason relationships of the detected instruments and tissues for pairing and determining actions.
For method evaluation, we construct a cataract surgery video dataset named PhacoQ and a cholecystectomy video dataset named CholecQ. Experiments on the PhacoQ Dataset and the CholecQ Dataset demonstrated that the proposed model outperforms existing state-of-the-art methods in the instrument-tissue interaction detection task with a large performance gain. We also provide a detailed study to prove the effectiveness of the proposed key components: SCF Layer, SCA Layer, and TG Layer. Further research is needed on the joint training and optimization of the two-stage model so that the overall interaction detection performance will be less influenced by the false positive detection of the first stage. Besides, other models such as Temporal Convolution Network (TCN) or Transformer for temporal information modeling will be explored in the future.

\bibliographystyle{IEEEtran}
\bibliography{ref}

\begin{thebibliography}{10}
\providecommand{\url}[1]{#1}
\csname url@samestyle\endcsname
\providecommand{\newblock}{\relax}
\providecommand{\bibinfo}[2]{#2}
\providecommand{\BIBentrySTDinterwordspacing}{\spaceskip=0pt\relax}
\providecommand{\BIBentryALTinterwordstretchfactor}{4}
\providecommand{\BIBentryALTinterwordspacing}{\spaceskip=\fontdimen2\font plus
\BIBentryALTinterwordstretchfactor\fontdimen3\font minus \fontdimen4\font\relax}
\providecommand{\BIBforeignlanguage}[2]{{%
\expandafter\ifx\csname l@#1\endcsname\relax
\typeout{** WARNING: IEEEtran.bst: No hyphenation pattern has been}%
\typeout{** loaded for the language `#1'. Using the pattern for}%
\typeout{** the default language instead.}%
\else
\language=\csname l@#1\endcsname
\fi
#2}}
\providecommand{\BIBdecl}{\relax}
\BIBdecl

\bibitem{maie2017surgical}
L.~Maier-Hein, S.~S. Vedula, S.~Speidel, N.~Navab, R.~Kikinis, A.~Park, M.~Eisenmann, H.~Feussner, G.~Forestier, S.~Giannarou \emph{et~al.}, ``Surgical data science for next-generation interventions,'' \emph{Nature Biomedical Engineering}, vol.~1, no.~9, pp. 691--696, 2017.

\bibitem{lin2022instrument}
W.~Lin, Y.~Hu, L.~Hao, D.~Zhou, M.~Yang, H.~Fu, C.~Chui, and J.~Liu, ``Instrument-tissue interaction quintuple detection in surgery videos,'' in \emph{International Conference on Medical Image Computing and Computer-Assisted Intervention}.\hskip 1em plus 0.5em minus 0.4em\relax Springer, 2022, pp. 399--409.

\bibitem{feng2009application}
X.~B. Feng and D.~H. Fu, ``Application of computer assisted navigation system in orthopaedics,'' \emph{Journal of Clinical Rehabilitative Tissue Engineering Research}, vol.~13, no.~30, pp. 5935--5938, 2009.

\bibitem{nwoye2020recognition}
C.~I. Nwoye, C.~Gonzalez, T.~Yu, P.~Mascagni, D.~Mutter, J.~Marescaux, and N.~Padoy, ``Recognition of instrument-tissue interactions in endoscopic videos via action triplets,'' in \emph{International Conference on Medical Image Computing and Computer-Assisted Intervention}.\hskip 1em plus 0.5em minus 0.4em\relax Springer, 2020, pp. 364--374.

\bibitem{islam2020learning}
M.~Islam, L.~Seenivasan, L.~C. Ming, and H.~Ren, ``Learning and reasoning with the graph structure representation in robotic surgery,'' in \emph{International Conference on Medical Image Computing and Computer-Assisted Intervention}.\hskip 1em plus 0.5em minus 0.4em\relax Springer, 2020, pp. 627--636.

\bibitem{xu2021learning}
M.~Xu, M.~Islam, C.~M. Lim, and H.~Ren, ``Learning domain adaptation with model calibration for surgical report generation in robotic surgery,'' in \emph{2021 IEEE International Conference on Robotics and Automation (ICRA)}.\hskip 1em plus 0.5em minus 0.4em\relax IEEE, 2021, pp. 12\,350--12\,356.

\bibitem{seenivasan2022global}
L.~Seenivasan, S.~Mitheran, M.~Islam, and H.~Ren, ``Global-reasoned multi-task learning model for surgical scene understanding,'' \emph{IEEE Robotics and Automation Letters}, vol.~7, no.~2, pp. 3858--3865, 2022.

\bibitem{nwoye2022rendezvous}
C.~I. Nwoye, T.~Yu, C.~Gonzalez, B.~Seeliger, P.~Mascagni, D.~Mutter, J.~Marescaux, and N.~Padoy, ``Rendezvous: Attention mechanisms for the recognition of surgical action triplets in endoscopic videos,'' \emph{Medical Image Analysis}, vol.~78, p. 102433, 2022.

\bibitem{li2022sirnet}
L.~Li, X.~Li, S.~Ding, Z.~Fang, M.~Xu, H.~Ren, and S.~Yang, ``Sirnet: Fine-grained surgical interaction recognition,'' \emph{IEEE Robotics and Automation Letters}, vol.~7, no.~2, pp. 4212--4219, 2022.

\bibitem{sarikaya2017detection}
D.~Sarikaya, J.~J. Corso, and K.~A. Guru, ``Detection and localization of robotic tools in robot-assisted surgery videos using deep neural networks for region proposal and detection,'' \emph{IEEE Transactions on Medical Imaging}, vol.~36, no.~7, pp. 1542--1549, 2017.

\bibitem{lea2015an}
C.~Lea, G.~D. Hager, and R.~Vidal, ``An improved model for segmentation and recognition of fine-grained activities with application to surgical training tasks,'' in \emph{2015 IEEE Winter Conference on Applications of Computer Vision}, 2015, pp. 1123--1129.

\bibitem{khatibi2020proposing}
T.~Khatibi and P.~Dezyani, ``Proposing novel methods for gynecologic surgical action recognition on laparoscopic videos,'' \emph{Multimedia Tools and Applications}, vol.~79, no.~41, pp. 30\,111--30\,133, 2020.

\bibitem{lea2016temporal}
C.~Lea, R.~Vidal, A.~Reiter, and G.~D. Hager, ``Temporal convolutional networks: A unified approach to action segmentation,'' in \emph{European conference on computer vision}.\hskip 1em plus 0.5em minus 0.4em\relax Springer, 2016, pp. 47--54.

\bibitem{dipietro2016recognizing}
R.~DiPietro, C.~Lea, A.~Malpani, N.~Ahmidi, S.~S. Vedula, G.~I. Lee, M.~R. Lee, and G.~D. Hager, ``Recognizing surgical activities with recurrent neural networks,'' in \emph{International conference on medical image computing and computer-assisted intervention}.\hskip 1em plus 0.5em minus 0.4em\relax Springer, 2016, pp. 551--558.

\bibitem{bawa2021saras}
V.~S. Bawa, G.~Singh, F.~KapingA, I.~Skarga-Bandurova, E.~Oleari, A.~Leporini, C.~Landolfo, P.~Zhao, X.~Xiang, G.~Luo, K.~Wang, L.~Li, B.~Wang, S.~Zhao, L.~Li, A.~Stabile, F.~Setti, R.~Muradore, and F.~Cuzzolin, ``The saras endoscopic surgeon action detection (esad) dataset: Challenges and methods,'' \emph{ArXiv}, vol. abs/2104.03178, 2021.

\bibitem{qi2018learning}
S.~Qi, W.~Wang, B.~Jia, J.~Shen, and S.-C. Zhu, ``Learning human-object interactions by graph parsing neural networks,'' in \emph{Proceedings of the European conference on computer vision (ECCV)}, 2018, pp. 401--417.

\bibitem{hamilton2017inductive}
W.~Hamilton, Z.~Ying, and J.~Leskovec, ``Inductive representation learning on large graphs,'' \emph{Advances in neural information processing systems}, vol.~30, 2017.

\bibitem{chao2015hico}
Y.-W. Chao, Z.~Wang, Y.~He, J.~Wang, and J.~Deng, ``Hico: A benchmark for recognizing human-object interactions in images,'' in \emph{Proceedings of the IEEE international conference on computer vision}, 2015, pp. 1017--1025.

\bibitem{vaswani2017attention}
A.~Vaswani, N.~Shazeer, N.~Parmar, J.~Uszkoreit, L.~Jones, A.~N. Gomez, {\L}.~Kaiser, and I.~Polosukhin, ``Attention is all you need,'' \emph{Advances in neural information processing systems}, vol.~30, 2017.

\bibitem{tamura2021qpic}
M.~Tamura, H.~Ohashi, and T.~Yoshinaga, ``Qpic: Query-based pairwise human-object interaction detection with image-wide contextual information,'' in \emph{2021 IEEE/CVF Conference on Computer Vision and Pattern Recognition (CVPR)}, 2021, pp. 10\,405--10\,414.

\bibitem{zou2021end}
C.~Zou, B.~Wang, Y.~Hu, J.~Liu, Q.~Wu, Y.~Zhao, B.~Li, C.~Zhang, C.~Zhang, Y.~Wei, and J.~Sun, ``End-to-end human object interaction detection with hoi transformer,'' in \emph{2021 IEEE/CVF Conference on Computer Vision and Pattern Recognition (CVPR)}, 2021, pp. 11\,820--11\,829.

\bibitem{zhang2022efficient}
F.~Z. Zhang, D.~Campbell, and S.~Gould, ``Efficient two-stage detection of human-object interactions with a novel unary-pairwise transformer,'' in \emph{Proceedings of the IEEE/CVF Conference on Computer Vision and Pattern Recognition}, 2022, pp. 20\,104--20\,112.

\bibitem{maier2021heidelberg}
L.~Maier-Hein, M.~Wagner, T.~Ross, A.~Reinke, and B.~P. Müller-Stich, ``Heidelberg colorectal data set for surgical data science in the sensor operating room,'' \emph{Scientific Data}, vol.~8, no.~1, p. 101, 2021.

\bibitem{twinanda2017endonet}
A.~P. Twinanda, S.~Shehata, D.~Mutter, J.~Marescaux, M.~de~Mathelin, and N.~Padoy, ``Endonet: A deep architecture for recognition tasks on laparoscopic videos,'' \emph{IEEE Transactions on Medical Imaging}, vol.~36, no.~1, pp. 86--97, 2017.

\bibitem{arnaud2021micro}
A.~Huaulm{\'e}, D.~Sarikaya, K.~Le~Mut, F.~Despinoy, Y.~Long, Q.~Dou, C.-B. Chng, W.~Lin, S.~Kondo, L.~Bravo-S{\'a}nchez \emph{et~al.}, ``Micro-surgical anastomose workflow recognition challenge report,'' \emph{Computer Methods and Programs in Biomedicine}, vol. 212, p. 106452, 2021.

\bibitem{sanat2021multi}
S.~Ramesh, D.~Dall’Alba, C.~Gonzalez, T.~Yu, P.~Mascagni, D.~Mutter, J.~Marescaux, P.~Fiorini, and N.~Padoy, ``Multi-task temporal convolutional networks for joint recognition of surgical phases and steps in gastric bypass procedures,'' \emph{International Journal of Computer Assisted Radiology and Surgery}, vol.~16, no.~7, pp. 1111--1119, Jul. 2021.

\bibitem{valderrama2022towards}
N.~Valderrama, P.~Ruiz~Puentes, I.~Hern{\'a}ndez, N.~Ayobi, M.~Verlyck, J.~Santander, J.~Caicedo, N.~Fern{\'a}ndez, and P.~Arbel{\'a}ez, ``Towards holistic surgical scene understanding,'' in \emph{Medical Image Computing and Computer Assisted Intervention -- MICCAI 2022}, L.~Wang, Q.~Dou, P.~T. Fletcher, S.~Speidel, and S.~Li, Eds.\hskip 1em plus 0.5em minus 0.4em\relax Cham: Springer Nature Switzerland, 2022, pp. 442--452.

\bibitem{vivek2020esad}
V.~S. Bawa, G.~Singh, F.~KapingA, InnaSkarga-Bandurova, A.~Leporini, C.~Landolfo, A.~Stabile, F.~Setti, R.~Muradore, E.~Oleari, and F.~Cuzzolin, ``Esad: Endoscopic surgeon action detection dataset,'' \emph{ArXiv}, vol. abs/2006.07164, 2020.

\bibitem{zhao2022MURPHYRM}
S.~Zhao, Y.~Liu, Q.~Wang, D.~Sun, R.~Liu, and S.~Zhou, ``Murphy: Relations matter in surgical workflow analysis,'' \emph{ArXiv}, vol. abs/2212.12719, 2022.

\bibitem{hong2020cholecseg8k}
W.-Y. Hong, C.-L. Kao, Y.-H. Kuo, J.-R. Wang, W.-L. Chang, and C.-S. Shih, ``Cholecseg8k: a semantic segmentation dataset for laparoscopic cholecystectomy based on cholec80,'' \emph{arXiv preprint arXiv:2012.12453}, 2020.

\bibitem{ren2015faster}
S.~Ren, K.~He, R.~Girshick, and J.~Sun, ``Faster r-cnn: Towards real-time object detection with region proposal networks,'' \emph{Advances in neural information processing systems}, vol.~28, 2015.

\bibitem{lin2020focal}
T.-Y. Lin, P.~Goyal, R.~Girshick, K.~He, and P.~Dollár, ``Focal loss for dense object detection,'' \emph{IEEE Transactions on Pattern Analysis and Machine Intelligence}, vol.~42, no.~2, pp. 318--327, 2020.

\bibitem{gupta2015visual}
S.~Gupta and J.~Malik, ``Visual semantic role labeling,'' \emph{ArXiv}, vol. abs/1505.04474, 2015.

\bibitem{gao2018ican}
C.~Gao, Y.~Zou, and J.-B. Huang, ``ican: Instance-centric attention network for human-object interaction detection,'' in \emph{British Machine Vision Conference}, 2018.

\bibitem{zhang2021spatially}
F.~Z. Zhang, D.~Campbell, and S.~Gould, ``Spatially conditioned graphs for detecting human-object interactions,'' in \emph{Proceedings of the IEEE/CVF International Conference on Computer Vision}, 2021, pp. 13\,319--13\,327.

\bibitem{demvsar2006statistical}
J.~Dem{\v{s}}ar, ``Statistical comparisons of classifiers over multiple data sets,'' \emph{The Journal of Machine learning research}, vol.~7, pp. 1--30, 2006.

\end{thebibliography}

\end{document}